\relax
\documentclass[letterpaper]{article} 
\usepackage{aaai22}  
\usepackage{times}  
\usepackage{helvet} 
\usepackage{courier}  
\usepackage[hyphens]{url}  
\usepackage{graphicx} 
\usepackage{natbib}  
\usepackage{caption} 
\usepackage{color} 

\usepackage{amsfonts}
\usepackage{amsmath}
\usepackage{amssymb}
\usepackage{multirow}

\newcommand{\yz}[1]{{\color{black} #1}}

\begin{document}

\title{Anomaly Discovery in Semantic Segmentation \\ via Distillation Comparison Networks}

\author{
Huan Zhou$^{1,\#}$\quad 
Shi Gong$^{1,\#}$\quad 
Yu Zhou$^{1,\dagger}$ \quad 
Zengqiang Zheng$^{2}$ \quad
Ronghua Liu$^{2}$ \quad
Xiang Bai$^{1}$ \\
$^{1}$ Huazhong University of Science and Technology \\ \qquad
$^{2}$ Wuhan Jingce Electronic Group Co.,Ltd.\\
{\tt\small
\{huanzhou, gongshi, yuzhou, xbai\}@hust.edu.cn \{zhengzengqiang, liuronghua\}@wuhanjingce.com}
}

\maketitle

{\let\thefootnote\relax\footnote{\# Equal contribution.}}
{\let\thefootnote\relax\footnote{$\dagger$ Corresponding author.}}

\begin{abstract}

This paper aims to address the problem of anomaly discovery in semantic segmentation.
Our key observation is that semantic classification plays a critical role in existing approaches, 
while the incorrectly classified pixels are easily regarded as anomalies.
Such a phenomenon frequently appears and is rarely discussed, 
which significantly reduces the performance of anomaly discovery.
To this end, we propose a novel Distillation Comparison Network (DiCNet).
It comprises of a teacher branch which is a semantic segmentation network that removed the semantic classification head, 
and a student branch that is distilled from the teacher branch through a distribution distillation. 
We show that the distillation guarantees the semantic features of the two branches hold consistency in the known classes, 
while reflect inconsistency in the unknown class.
Therefore, we leverage the semantic feature discrepancy between the two branches to discover the anomalies. 
DiCNet abandons the semantic classification head in the inference process, 
and hence significantly alleviates the issue caused by incorrect semantic classification. 
Extensive experimental results on StreetHazards dataset and BDD-Anomaly dataset are conducted to verify the superior performance of DiCNet.
In particular, DiCNet obtains a 6.3\% improvement in AUPR and a 5.2\% improvement in FPR95 on StreetHazards dataset,
achieves a 4.2\% improvement in AUPR and a 6.8\% improvement in FPR95 on BDD-Anomaly dataset.
Codes are available at \url{https://github.com/zhouhuan-hust/DiCNet.}

\end{abstract}

\section{Introduction}


Semantic segmentation \cite{Geng29, Yuan30} aims to assign a semantic class label for each pixel. 
These approaches are trained on a set of known semantic classes.
However, in the real world open scenario,
if a pixel belongs to an unknown class, i.e., the never-seen-before class,   
it is dangerous to classify it into a predefined class instead of anomaly one.
This paper focuses on exploiting Anomaly Discovery in Semantic Segmentation (ADSS),
which aims to discover the pixels belonging to the unknown class.
ADSS can be widely used in safety-critical applications, 
such as autonomous driving \cite{Janai27}, medical image analysis \cite{Liu28}.


\begin{figure}[t]
\centering
\includegraphics[width=1.0\columnwidth]{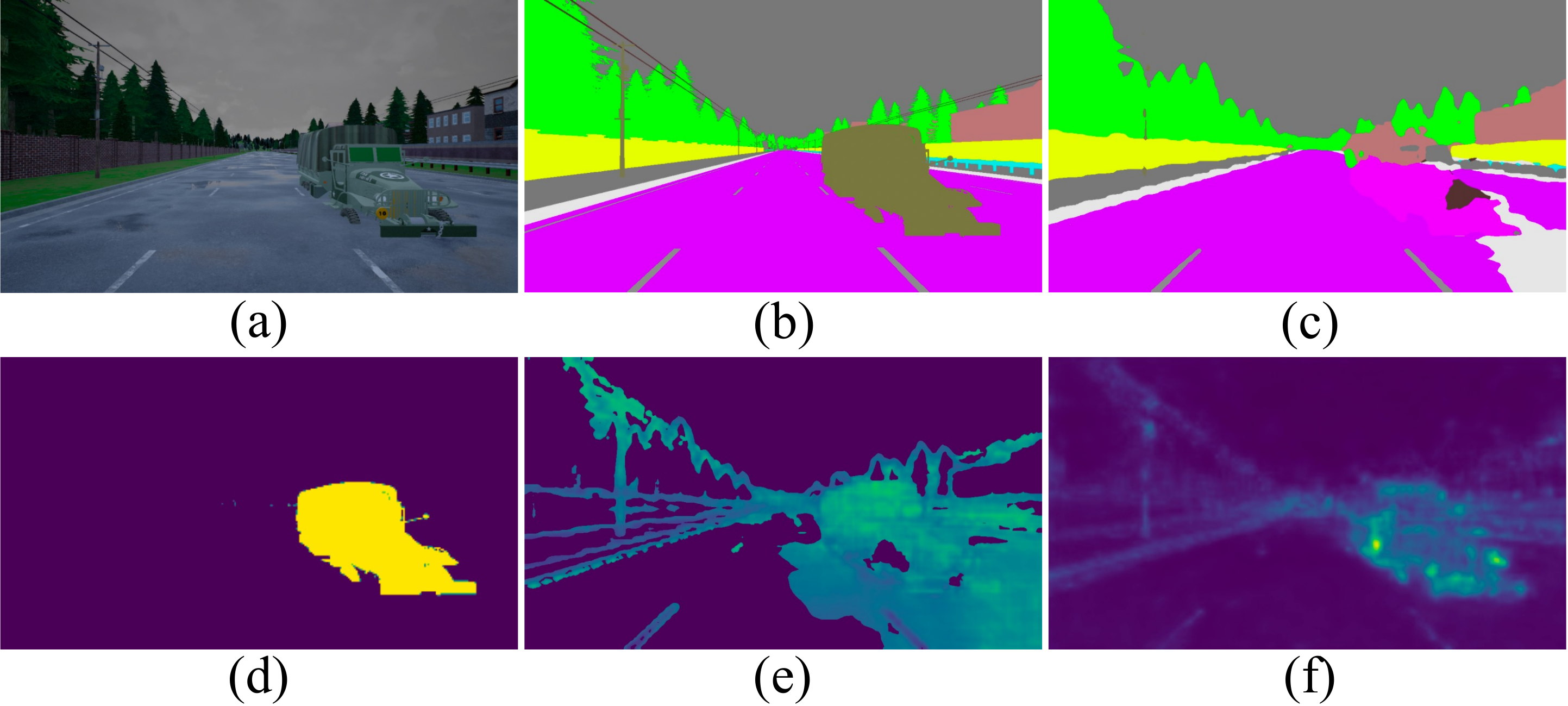} 
\caption{Comparison between SynthCP \cite{Xia03} and DiCNet. 
(a) is the input image, 
(b) is the ground truth of semantic segmentation, 
(c) is a semantic map predicted by PSPNet \cite{Zhao05}, 
(d) is the ground truth of anomaly, 
(e) is the anomaly score map of SynthCP, 
(f) is the anomaly score map of DiCNet. 
}
\label{fig:f1}
\end{figure}

Given the input image shown in Fig. \ref{fig:f1} (a),
existing approaches firstly leverage a semantic segmentation network, e.g., PSPNet \cite{Zhao05}, DeepLab \cite{chen51},  
to predict a semantic map, as shown in Fig. \ref{fig:f1} (c).
Then they use different strategies to predict the anomalies based on the predicted semantic map.
The uncertainty estimation based approaches \cite{Hendrycks19, Hendrycks02,Corbiere32}
conduct different post-processing on the predicted semantic map, e.g., Softmax \cite{Hendrycks19}, CRF \cite{Hendrycks02}, 
semantic map ensemble \cite{Franchi17,Gal10,Izmailov15,Lakshminarayanan13,Maddox16,Vyas14}, 
to discover the anomalies. 
The image re-synthesis based methods \cite{Lis01,Xia03,di50}
use the predicted semantic map to re-synthesis the original image,
and compare the re-synthesis image with the original one to discover the anomalies. 
Hence, it is relatively apparent that semantic prediction is a core step in existing ADSS approaches.

%

\begin{table}[!tb]
\centering
\begin{tabular}{l|c|c|c}
    \hline
	Method & E-FPR95$\downarrow$  & E-FPR85$\downarrow$  & E-FPR75$\downarrow$       \\  
	\hline
	SynthCP   &85.2  & 72.8  &62.3     \\
	DiCNet   &61.7  & 48.0  &39.5      \\
	\hline
	\hline
	Method & C-FPR95$\downarrow$  & C-FPR85$\downarrow$  & C-FPR75$\downarrow$      \\  
	\hline
	SynthCP   & 23.0  &15.1  &11.3    \\
	DiCNet    & 13.4  & 8.6  &6.3 \\
	  		
	\hline
\end{tabular}
\caption{
We decompose the normal pixels into two subsets, 
namely, the normal pixels that are correctly classified, 
and the normal pixels that are incorrectly classified.
E-FPR95 measures how many incorrectly classified pixels are predicted as anomalies when the recall is 95\%.
C-FPR95 reflects how many correctly classified pixels are predicted as anomalies when the recall is 95\%.
We give the results of E-FPR and C-FPR at different recalls.
}
\label{tab:efpr}
\end{table}

\textbf{Our Finding:}
\emph{Since most of the existing approaches heavily depend on the semantic segmentation results,
we observe that the incorrectly classified pixels in semantic segmentation are easily regarded as anomalies.
For instance, when compared with the ground truth given in Fig. \ref{fig:f1} (b),
the predicted semantic map in Fig. \ref{fig:f1} (c) contains some incorrectly classified pixels, 
e.g., the pixels near the edge of the tree, the pixels inside the lower right corner region of the road. 
Accordingly, these pixels are predicted as anomalies by the existing state-of-the-art approach, i.e., SynthCP \cite{Xia03}, 
as seen in Fig. \ref{fig:f1} (e).
The additional statistical analysis is reported in Tab. \ref{tab:efpr}. 
For SynthCP, 85.2\% of the incorrectly classified pixels are predicted 
as anomalies when the recall is 95\% (E-FPR95), 
and the consistent conclusion can be observed at different recalls.
Such a phenomenon seriously influences the accuracy of the anomaly discovery, 
while it is rarely discussed in existing approaches.
}

In this paper, we propose a Distillation Comparison Network (DiCNet) for anomaly discovery in semantic segmentation, which includes a teacher branch and a student branch.
Both the teacher branch and the student branch are formed as a semantic segmentation network that removed the semantic classification head.
The training phase of DiCNet includes two steps.
The teacher branch is firstly trained under the common setting of the semantic segmentation,
and then the student branch is distilled from the fixed teacher branch by the distribution distillation.
The goal of the distillation is to guarantee that the teacher branch and the student branch hold consistency in the known classes,
while \yz{reflect inconsistency} in the unknown class.
In the testing phase, we directly compare the output features of the teacher branch and the student branch.
The discrepancy of these two branches is employed to assign an anomaly score for each pixel.
As shown in Fig. \ref{fig:f1} (f), 
DiCNet alleviates the issue caused by incorrect semantic classification efficiently.
In addition, as shown in Tab. \ref{tab:efpr}, 
DiCNet significantly reduces the E-FPR at different recalls.
Furthermore, DiCNet also greatly reduces the C-FPR, 
which indicates that DiCNet is capable of produce fewer  errors in the correctly classified pixels than SynthCP.
We validate DiCNet on two challenging datasets, i.e., StreetHazards dataset and BDD-Anomaly dataset \cite{Hendrycks02} and show its superiority to existing ADSS approaches. 
To be specific, DiCNet achieves improvements over the state-of-the-art by $6.3\%$ AUPR and $5.2\%$ FPR95 on StreetHazards dataset,
$4.2\%$ AUPR and $6.8\%$ FPR95 on BDD-Anomaly dataset.

To wrap up, the main contributions of this work lie in the following aspects:
\begin{itemize}
    \item We give a deep study of the ADSS task, 
    and reveal that the incorrectly classified pixels in semantic prediction are easily regarded as anomalies, 
    leading to poor accuracy of ADSS. 
    To the best of our knowledge, such an essential problem is rarely discussed in existing approaches.
    \item We customize a Distillation Comparison Network (DiCNet) 
    to deal with the issue caused by the incorrectly classified pixels.
    DiCNet alleviates such a problem efficiently,
    and gains a significant improvement on two challenging datasets.
    Furthermore, DiCNet is simple and flexible.

\end{itemize}


\section{Related Work}

In this section, we briefly review the approaches about anomaly discovery in semantic segmentation and knowledge distillation, 
which are closely related to our approach. 

\subsection{Anomaly Discovery in Semantic Segmentation}

Existing anomaly discovery in semantic segmentation approaches can be roughly classified into two categories,
i.e., uncertainty estimation based approaches \cite{Geifman46,Jiang47,Vyas14,fontanel49} and 
image re-synthesis based approaches \cite{Lis01,Xia03,di50}.

\noindent\textbf{Uncertainty Estimation based Approaches.}
These approaches first acquire a semantic prediction map by using a semantic segmentation network.
\cite{Hendrycks19} proposes the maximum softmax probability (MSP) to estimate the uncertainty of the semantic prediction,
and the pixel with a higher uncertainty is regarded as an anomaly. 
However, as argued in \cite{Corbiere32}, 
the existing deep networks tend to produce a lower uncertainty for most of the pixels, which reduce the effectiveness of MSP.
Hence \cite{Corbiere32} proposes a true class probability for anomaly discovery, which improves the accuracy of MSP.
In addition, recent approaches \cite{Izmailov15,Maddox16} train multiple semantic segmentation networks, 
and estimate the uncertainty according to the variance of different semantic segmentation networks.  
\cite{Gal10} proposes the Monte Carlo Dropout to produce a set of semantic prediction results,  
but \cite{Hendrycks02,Lis01} argue that MC Dropout fails to detect anomalies and produces a lot of false positives. 
\cite{Lakshminarayanan13} proposes the ensemble models trained independently with different initialization seeds.
\cite{Franchi17} \yz{proposes to} compute the weight distributions, which can be used to sample an ensemble of networks for estimating the uncertainty. 

\noindent\textbf{Image Re-synthesis based Approaches.} 
\cite{Akcay07,Baur11,Creusot08,Munawar09} use autoencoder to segment the anomalies.
They assume that the never-seen-before objects cannot be decoded accurately. 
However, autoencoder tends to generate a lower-quality image.
For the complex scenes with high variability in scene layout and lighting, 
they fail to find anomaly objects. 
\cite{Lis01,Xia03,di50} leverage a segmentation network \cite{Badrinarayanan04,Zhao05} to obtain the semantic map, 
then the generative network \cite{Schlegl12,Wang06}, such as GAN, is employed to learn a mapping from semantic map to the re-synthesis image. 
Since the re-synthesis image only contains the \yz{normal classes, 
the anomaly class} can be discovered by comparing the re-synthesis image and the original one. 

As we previously discussed, the semantic segmentation results \yz{exert a great impact on} anomaly discovery in existing approaches, 
since the incorrectly classified pixels are easily regarded as anomalies.
DiCNet alleviates this problem \yz{by comparing the discrepancy in feature space,}
which avoids errors caused by the semantic classification process.

\subsection{Knowledge Distillation}

Knowledge distillation aims to transfer the knowledge of a large network (a.k.a teacher network) to a small network (a.k.a student network).
\yz{The student network may have an approximate predictive ability as the teacher network.}
Knowledge distillation methods can be divided into three types: 
logits-based approaches \cite{Cho36,Hinton34,Phuong39,Xie37,Yang38,Zhang35}, 
feature-based approaches \cite{Komodakis41,Romero40}, 
and relation-based approaches \cite{Tung43,Yim42,Romero40}. 
In this paper, we follow the feature-based paradigm \cite{Romero40} by proposing a distribution distillation.
However, DiCNet is flexible since there are no strict constraints for the model size of the student branch.


\section{Approach}

In this section, 
we first formalize the anomaly discovery in semantic segmentation problem in Sec. \ref{problem_formulation},
and then describe the specific network architecture in Sec. \ref{network_architecture}. 
The distribution distillation and the scoring rule are introduced in Sec. \ref{sec:distillation} and Sec. \ref{scoring_rule}, respectively.

\subsection{Problem Formulation}
\label{problem_formulation}

\begin{figure}[t]
\centering
\includegraphics[width=1.0\columnwidth]{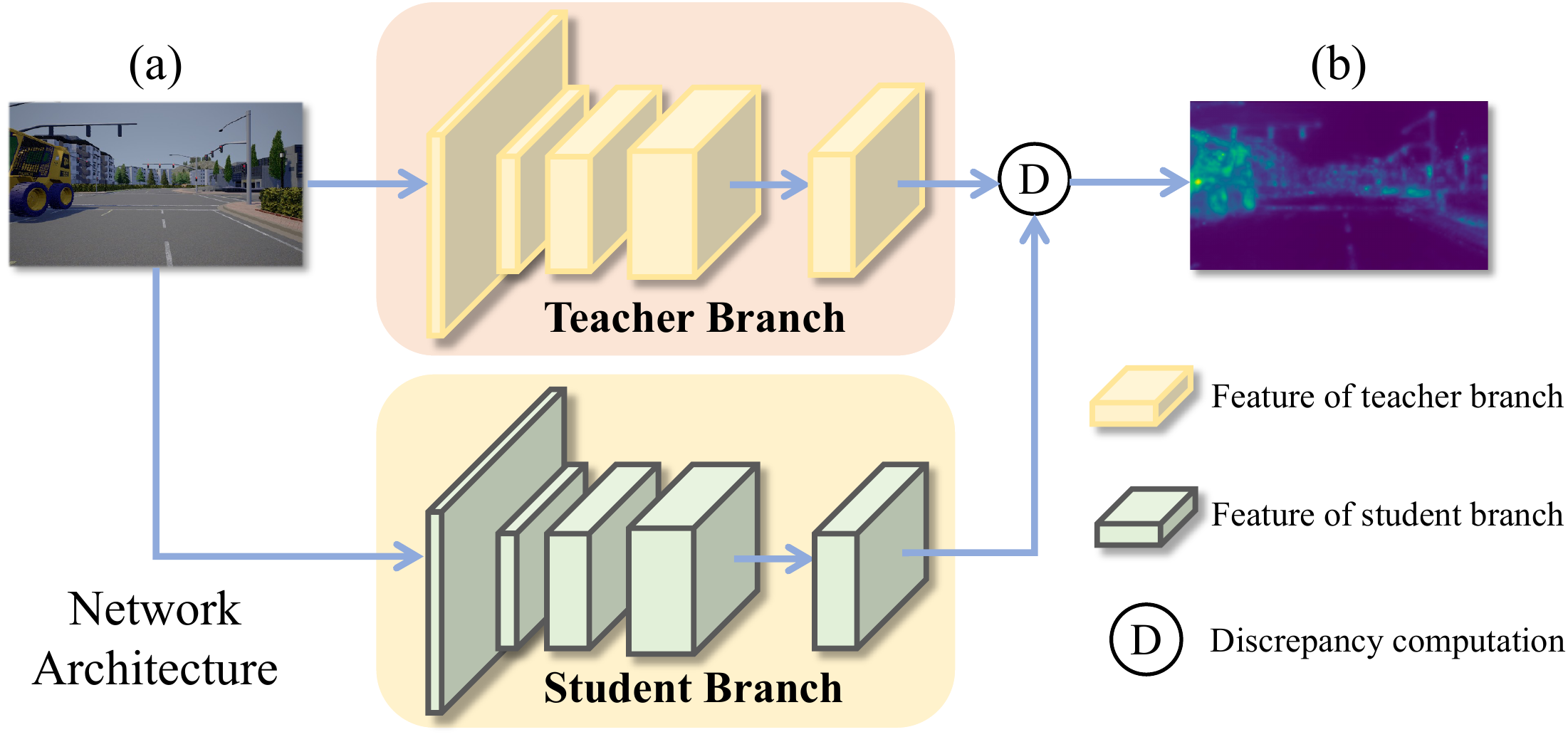} 
\caption{The network architecture of DiCNet. (a) is the input image. 
DiCNet contains a teacher branch and a student branch. 
The anomaly score is obtained by calculating the discrepancy between the two branches.
(b) is the anomaly score map predicted by DiCNet.}
\label{fig:arch}
\end{figure}

Given an input RGB image $\textit{I} \in \mathbb{R}^{H \times W \times 3}$, 
the goal of anomaly discovery in semantic segmentation is to predict an anomaly score map $\textit{A}\in \mathbb{R}^{H \times W \times 1}$.
Each element of $\textit{A}$ indicates the probability \yz{that it belongs} to the unknown class, 
i.e., the never-seen-before class.
In the training phase, the pixel-level annotations of image $I$ are given, 
and each pixel of $I$ belongs to one of the $\mathcal{Z}$ semantic classes.
The testing set contains $\mathcal{Z}+1$ classes, 
and the pixels that belong to the additional unknown class are regarded as anomalies. They are assigned a high probability in $A$. 

In this paper, we present a novel Distillation Comparison Network (DiCNet) for anomaly discovery in semantic segmentation.
As illustrated in Fig. \ref{fig:arch},
the overall structure of DiCNet includes a teacher branch and a student branch (see Sec. \ref{network_architecture}).
Both the teacher branch and the student branch are formed as a semantic segmentation network that removed the semantic classification head. 
The training phase of DiCNet is divided into two steps.
The teacher branch is first trained under the common setting of the semantic segmentation,
and then the student branch is distilled from the fixed teacher branch
by using the distribution distillation (see Sec. \ref{sec:distillation}). 
During the testing phase, 
the anomaly score of each pixel is assigned by the discrepancy of these two branches.

\subsection{Network Architecture}
\label{network_architecture}

\begin{figure}[t]
\centering
\includegraphics[width=1.0\columnwidth]{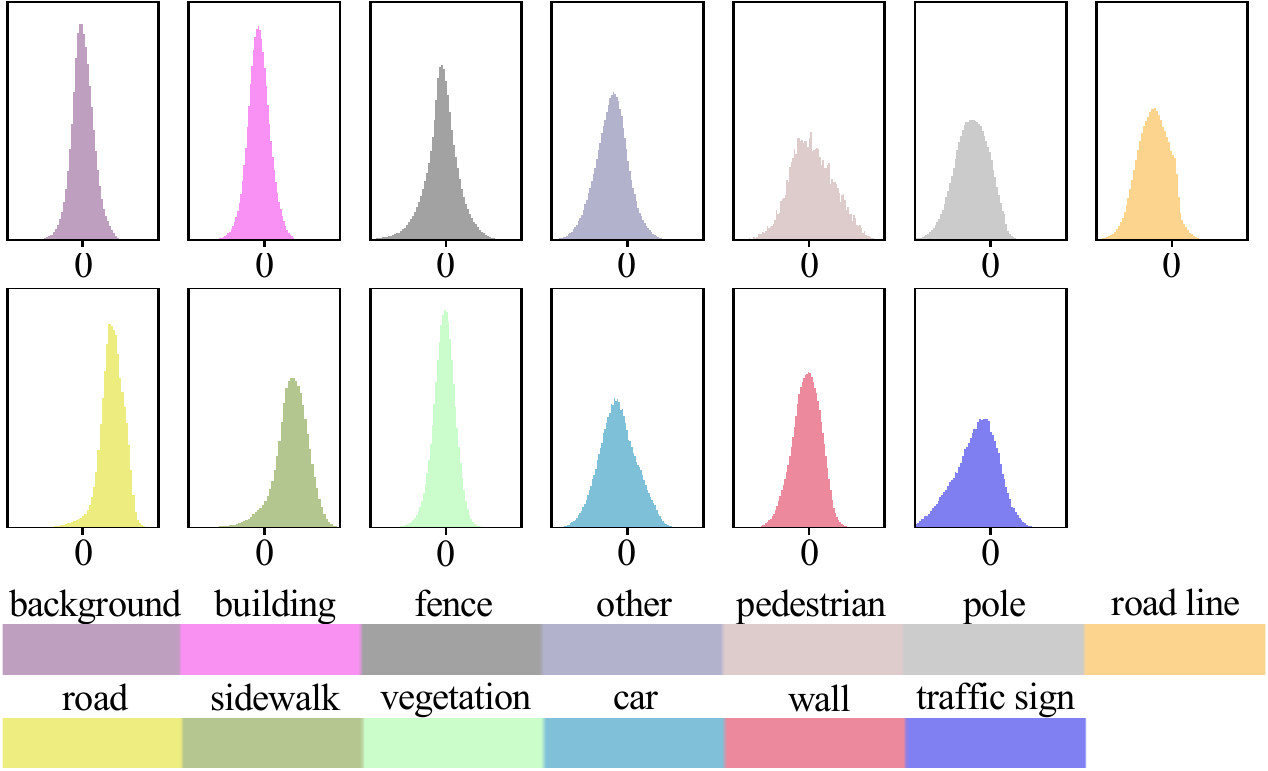} 
\caption{Numerical distribution of semantic features extracted from the training set of StreetHazards.}
\label{fig:dis}
\end{figure}

\noindent\textbf{Teacher Branch.}
In DiCNet, any semantic segmentation approach can be employed as the teacher branch, 
and the teacher branch is formulated as:
\begin{equation}
    \textbf{f}_t = \mathcal{U}(f_t) = \mathcal{U}(\theta_t(I)), 
\label{eq:teacher}
\end{equation}
where $\theta_t$ indicates the model parameter of the teacher branch, 
and $f_t\in \mathbb{R}^{\frac{H}{8} \times \frac{W}{8} \times C}$ is the semantic features.
Following \cite{Xia03}, the ResNet-50 based PSPNet \cite{Zhao05} is used as the teacher branch in this paper.
In addition, $\mathcal{U}$ denotes the channel-wise normalization. 
Let $f^{c}_{t} \in \mathbb{R}^{\frac{H}{8} \times \frac{W}{8} \times 1}$ denotes the $c$-th feature channel of $f_t$, where $c \in \{1,2,...,C\}$. 
Since the values of $f^{c}_{t}$ change in different ranges, 
it is not easy to distill its distribution. 
Therefore we calculate the mean $m^{c}_{t}$ and the standard variance $b^{c}_{t}$ of $f^{c}_{t}$, 
and then $f^{c}_{t}$ is normalized as follows,
\begin{equation}
\label{Eq:1}
\textbf{f}^{c}_{t} = \frac{f^{c}_{t} - m^{c}_{t}}{b^{c}_{t}}.
\end{equation}
Through the observations on StreetHazards dataset \cite{Hendrycks02},
we find that in each $\textbf{f}^{c}_{t}$,
the feature values of each semantic class fit a single Gaussian distribution,   
as shown in Fig. \ref{fig:dis}.
Consequently, $\textbf{f}^{c}_{t}$ fits a Gaussian mixture distribution.

\noindent\textbf{Student Branch.}
The student branch also receives $I$ as the input, 
let $\theta_s$ denotes the model parameter of the student branch, thus,
\begin{equation}
    \textbf{f}_s = \mathcal{U}(f_s) = \mathcal{U}(\theta_s(I)), 
\label{eq:student}
\end{equation}
where $\textbf{f}_s\in \mathbb{R}^{\frac{H}{8} \times \frac{W}{8} \times C}$ is the output of the student branch.
\yz{In the training phase, 
we want to optimize $\theta_s$ to guarantee $\textbf{f}_s$ and $\textbf{f}_t$ are as close as possible in the known $\mathcal{Z}$ classes.}
Thus, we design the student branch that \yz{has a similar network structure as the teacher branch.} 
In addition, different from existing knowledge distillation approaches \cite{Cho36,Hinton34,Phuong39,Xie37,Yang38,Zhang35,Komodakis41,Romero40},
which aim to distill the knowledge from a larger network to a smaller network. 
In our approach, the distillation is more flexible,
since we can distill the knowledge from a fixed teacher network to any model size, 
e.g., we can distill the knowledge of a ResNet-50 based teacher branch to a ResNet-101 based student branch.
We will experimentally demonstrate in Sec. \ref{sec:discussions} that, 
given a fixed teacher branch, 
DiCNet is \yz{less sensitive} to the model size of the student branch. 
The accuracy difference between using different ResNet backbones is within 1\%.




\subsection{Distribution Distillation}
\label{sec:distillation}

Since we hope that $\textbf{f}_{t}$ and $\textbf{f}_{s}$ satisfy the same distribution 
in the known $\mathcal{Z}$ classes,
we define the discrepancy variable $\textbf{D}=\textbf{f}_{t} \ominus \textbf{f}_{s}$, 
where $\ominus$ denotes the element-wise subtraction,
thus, $\textbf{D} \in \mathbb{R}^{\frac{H}{8} \times \frac{W}{8} \times C}$.
Based on Eq. \eqref{eq:teacher}, 
$\textbf{f}_{t}$ is the conditional distribution of $I$ and $\theta_{t}$, 
which can be denoted as $P(\textbf{f}_t| I, \theta_{t})$. 
$\textbf{f}_{s}$ follows the same rules.
Thus, 
\begin{equation}
P(\textbf{D}|\textbf{f}_{t}, \textbf{f}_{s})  = P(\textbf{D}|I, \theta_{t}, \theta_{s})  
\end{equation}
Therefore, each element $d^{c}_{i,j} \in \textbf{D}$ satisfies a Gaussian distribution,
i.e., $d^{c}_{i,j} \sim \mathcal{N}(\mu^{c},(\sigma^{c})^{2})$.
In the limit case, iff $\textbf{f}_{t} = \textbf{f}_{s}$, $\textbf{D} = 0$,
and each element $d^{c}_{i,j} \in \textbf{D}$ is equal to $0$.
In such condition, $\mu^{c} = 0$, $\sigma^{c} = 0$, and $d^{c}_{i,j}$ satisfies the Dirac delta function. 
Such a phenomenon demonstrates that the student branch predicts the same results as the teacher branch in the known $\mathcal{Z}$ classes. 
Therefore, the goal of the distribution distillation is to optimize the parameter $\theta_s$ to minimize $|\mu^{c}|$ and $\sigma^{c}$. 
Since we have the following equivalence relation, 
\begin{equation}
\mu^{c}=0, \sigma^{c}=0 \iff \textbf{D} = 0 \iff \textbf{f}_{t} = \textbf{f}_{s}
\end{equation}
we can minimize each element $d^{c}_{i,j} \in \textbf{D}$ as follows,
\begin{figure}[t]
\centering
\includegraphics[width=1.0\columnwidth]{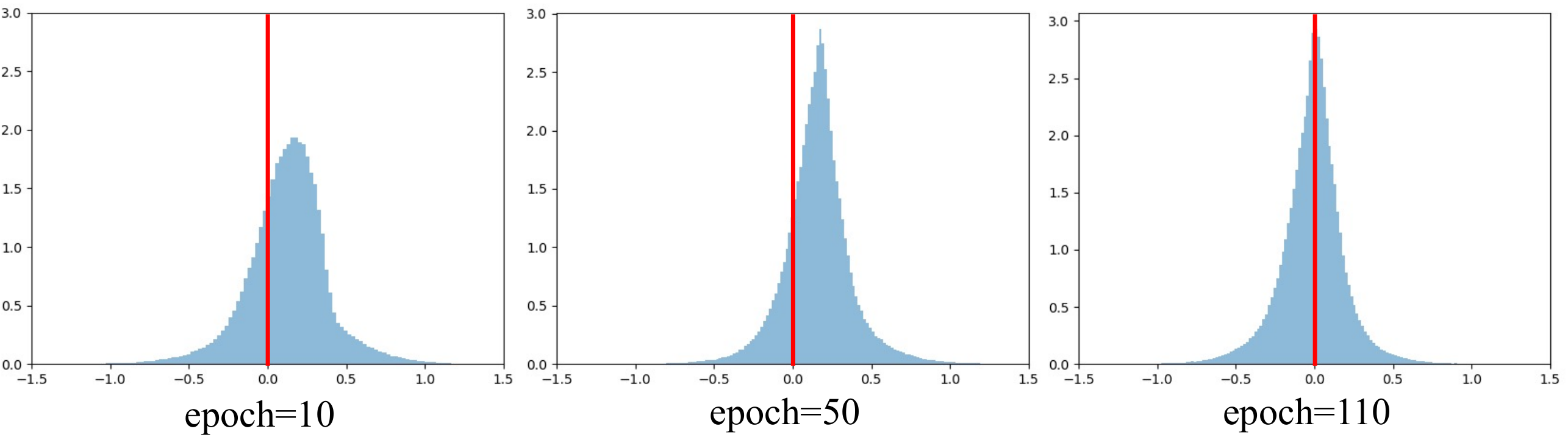} 
\caption{
The distribution variations of $d^{c}_{i,j}$ at different training epoch.
(The backbone of teacher branch is ResNet-50, and the backbone of student branch is ResNet-34).
}
\label{fig:d}
\end{figure}
\begin{equation}
\label{Eq:6}
\mathcal{L} = \frac{1}{M \times C}\sum\nolimits_{i,j,c}(d^{c}_{i,j})^2.
\end{equation}
where $M$ denotes the mini-batch size.
Since reducing $d^{c}_{i,j}$ is equivalent to reduce $\sigma^{c}$ and move $\mu^{c}$ to $0$,
the distribution of $d^{c}_{i,j}$ is inconsistent at different epoch.
As shown in Fig. \ref{fig:d}, 
when using a ResNet-50 based teacher branch, 
and a ResNet-34 based student branch. 
At epoch 10, the mean $\mu^{c}$ deviates from $0$, 
and the standard deviation $\sigma^{c}$ is large.
At epoch 50, the standard deviation $\sigma^{c}$ becomes smaller, 
while the mean $\mu^{c}$ still deviates from $0$.
At epoch 110, the mean $\mu^{c}$ is close to $0$.
We should emphasize that the optimization process \yz{cannot guarantee that $\mu^{c}$ achieves   the limitation case},  
i.e., $\mu^{c}$ is strictly equal to $0$. 
Hence, each channel may have an individual $\mu^{c}$. 
In addition, we observe that, by using a fixed teacher branch, 
the convergence speed of the distillation is related to the selection of the backbone.
A larger backbone may lead to a faster convergence, 
while a smaller backbone needs more training epochs.
However, it is fairly straightforward that a larger backbone may increase the inference time.
Therefore, we can choose the best structure of DiCNet by jointly considering the accuracy, the convergence speed, and the inference time.
More detailed results and discussions are given in Sec. \ref{sec:discussions}.

\subsection{Scoring Rule}
\label{scoring_rule}

\begin{figure}[t]
\centering
\includegraphics[width=1.0\columnwidth]{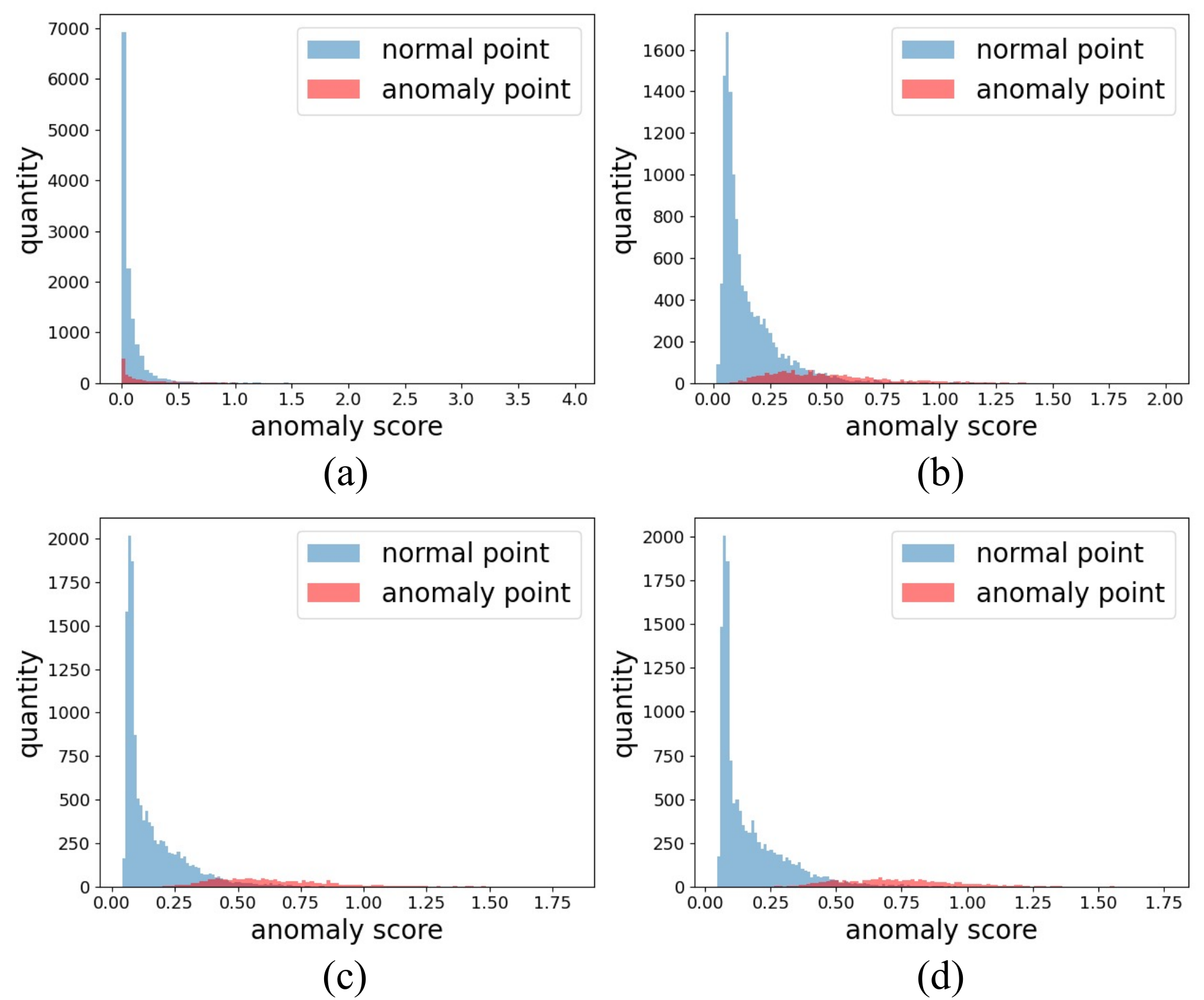} 
\caption{Anomaly score distribution with different channel number. (a) $C = 1$, (b) $C = 10$,(c) $C = 100$, (d) $C = 512$.}
\label{fig:score}
\end{figure}

Given a test image $I$, we input it into both the teacher branch and the student branch, 
and compute the discrepancy $\textbf{D}$.  
As we discussed in Sec. \ref{sec:distillation}, through the distribution distillation process, 
each element $d_{i,j}^c \in \textbf{D}$ satisfies a Gaussian distribution in the known $\mathcal{Z}$ classes,
while obeys a random distribution in the unknown class.
Based on this observation, we define the anomaly score $a_{i,j} \in \tilde{A}$ as the channel-wise variance of $d_{i,j}^c$: 
\begin{equation}
\begin{aligned}
\label{Eq:Scoring}
a_{i,j} 
& = \frac{1}{C} \sum\nolimits_{c}(d_{i,j}^c - \mu^{c} )^2 \\
\end{aligned}
\end{equation}
where $\tilde{A} \in \mathbb{R}^{\frac{H}{8} \times \frac{W}{8} \times 1} $.
The intuitive motivation of Eq. \eqref{Eq:Scoring} is that, 
if each $d_{i,j}^c \sim \mathcal{N}(\mu^{c}, (\sigma^{c})^{2})$, 
$d_{i,j}^c$ may be close to $\mu^{c}$ in the known $\mathcal{Z}$ classes,
and hence, $a_{i,j}$ may get a small anomaly score.
In contrast, if $d_{i,j}^c$ presents an arbitrary value that does not obey the learned Gaussian distribution,
a larger score is assigned to $a_{i,j}$ according to Eq. \eqref{Eq:Scoring},
which is regarded as an anomaly in this condition.
In addition, in most cases, 
the Gaussian distribution and the random distribution are inseparable in a single channel. 
Thus, channel number $C$ plays an important role in separating the anomaly from the normal regions. 
As shown in Fig. \ref{fig:score}, 
when $C = 1$, the value of the anomaly points 
(red points in Fig. \ref{fig:score} (a)) and normal points (blue points in Fig. \ref{fig:score} (a)) are mixed together.
When we increase $C$ to 10, a few anomalies show separability in terms of the feature value. 
As the channel number continues to increase, the overlap between the normal scores and the anomalies scores becomes smaller, 
e.g., when $C=512$, most of the anomalies can be separated out. 
Finally, $\tilde{A}$ is upsampled 8 times to achieve the anomaly score map $A$.


\begin{figure*}[t]
\centering
\includegraphics[width=1.0\textwidth]{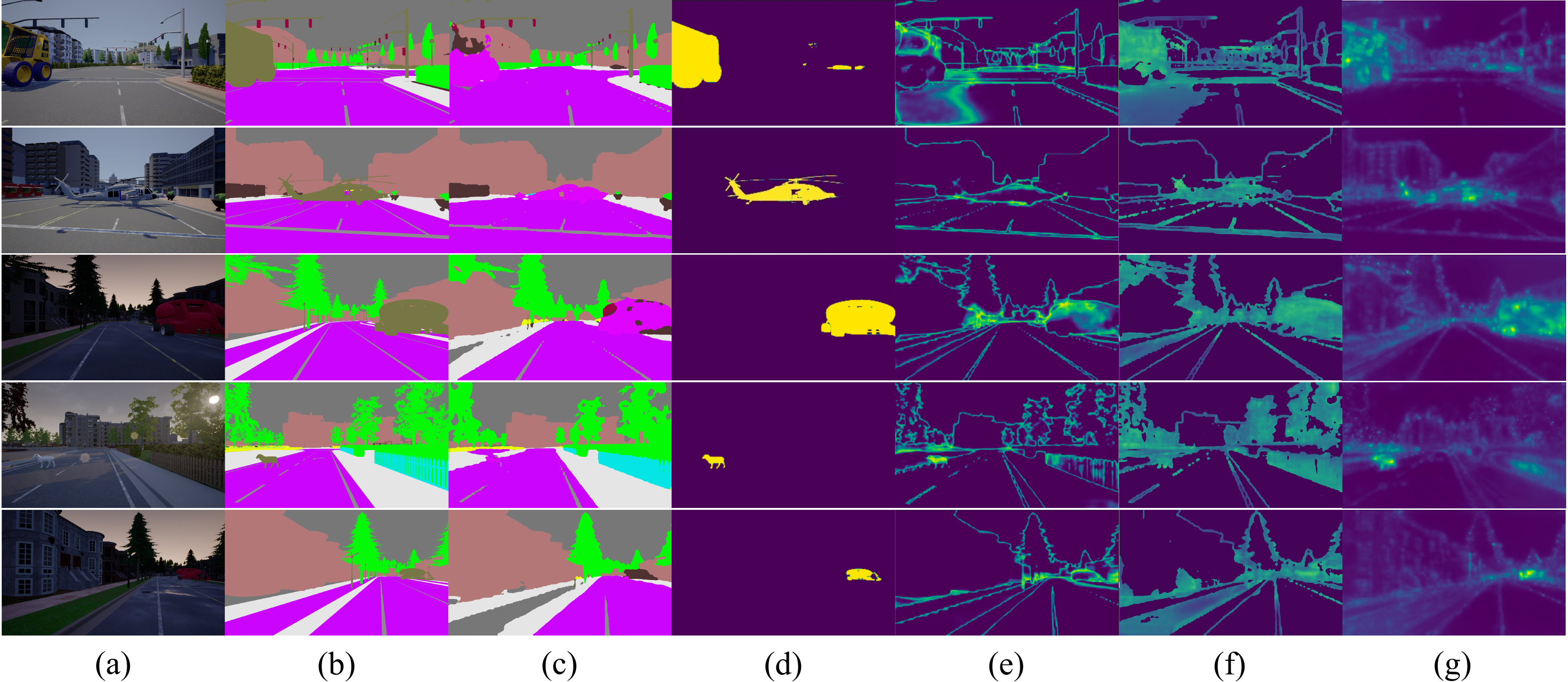} 
\caption{Qualitative comparisons with existing representative approaches. 
(a) the input images, 
(b) the ground truth of semantic segmentation, 
(c) the predicted semantic maps, 
(d) the ground truth of anomaly, 
(e) anomaly score maps of MSP \cite{Hendrycks19} , 
(f) anomaly score maps of SynthCP \cite{Xia03}, 
(g) anomaly score maps of DiCNet. 
Best view in color.}
\label{fig:compare}
\end{figure*} 

\section{Experiments}
\label{sec:experiments}

\subsection{Implementation Details}

\noindent\textbf{Model Training}.
DiCNet is implemented in PyTorch on Intel(R) Xeon(R) CPU and 4 NVIDIA TITAN Xp GPUs. 
Both the teacher branch and the student branch adopts a modified ResNet backbone pre-trained on ImageNet \cite{Deng22}. 
The results reported in Tab. \ref{tab:efpr}, Tab. \ref{tab:result} and Tab. \ref{tab:result_bdd} use ResNet-50 in teacher branch and ResNet-34 in student branch. 
More discussions about the selection of different backbones are given in Sec. \ref{sec:discussions}. 
For training the teacher branch, the mini-batch size is set to 2. 
For training the student branch, the mini-batch size is set to 4.
For StreetHazards dataset, We first train the teacher branch for 23 epochs with SGD using a learning rate of $2 \times 10^{-2}$, momentum of 0.9, and learning rate decay of $10^{-4}$.  Next, in the student branch training phase, the weights of teacher branch are frozen, student branch is trained for 110 epochs with Adam ($\beta_1$ = 0.9, $\beta_2$ = 0.999) optimizer. We use a learning rate of $10^{-4}$ and weight decay of $10^{-5}$.
For BDD-Anomaly dataset, We first train the teacher branch for 19 epochs with SGD with momentum using a learning rate of $2 \times 10^{-2}$, momentum of 0.9, and learning rate decay of $10^{-4}$. Next, in the student branch training phase, the weights of teacher branch are frozen, student branch is trained for 160 epochs with Adam ($\beta_1$ = 0.9, $\beta_2$ = 0.999) optimizer. The initial learning rate is set to $10^{-4}$ and the learning rate drops by 0.1 at the 65th epoch. The weight decay is set to $10^{-5}$.

\noindent\textbf{Datasets}.
StreetHazards \cite{Hendrycks02} leverages simulation to provide diverse, realistically-inserted anomaly objects, containing 5125 training images, 1031 validation images, and 1500 testing images. The image resolution is 1280 $\times$ 720. 
There are 250 types of anomaly objects appearing only in the testing images, 
and \yz{they are regarded as a single anomaly class.} 
BDD-Anomaly \cite{Hendrycks02} is a large-scale dataset with diverse driving conditions derived from the BDD100K semantic segmentation dataset \cite{Yu45}. The dataset contains 6,688 training images, 951 validation images without anomalies, and 361 testing images with anomalies. 
The image resolution is 1280 $\times$ 720. 
There are only two anomaly types, including motorcycle and train, and \yz{they are regarded as a single anomaly class}.

\noindent\textbf{Evaluation Metrics}. 
Following \cite{Hendrycks02,Hendrycks19}, we evaluate the performance of anomaly discovery
by three metrics: Area Under the ROC curve (AUROC), False Positive Rate at 95\% Recall (FPR95), Area Under the Precision-Recall curve (AUPR). 
Among these three metrics, we consider anomaly pixels to be positive and all the others to be negative.

\noindent\textbf{Baselines}.
We compare our method to MC Dropout \cite{Gal10}, MSP \cite{Hendrycks19}, Deep Ensemble \cite{Lakshminarayanan13}, AutoEncoder \cite{Baur11}, MSP+CRF \cite{Hendrycks02}, 
TRADI \cite{Franchi17}, SynthCP \cite{Xia03} and PAnS \cite{fontanel49}. 
The AutoEncoder method is firstly proposed for anomaly segmentation in medical images, which do not need semantic labels. All other methods are specifically designed for semantic segmentation.


\begin{table}[tb]
\small
\centering
    \begin{tabular}{l|c|c|c}
    \hline
	Method   & AUPR $\uparrow$ & FPR95 $\downarrow$  & AUROC $\uparrow$    \\  
	\hline
	\hline
	MC Dropout(2016)  & 7.5 & 79.4  & 69.9   \\
	MSP(2017)  & 6.6 & 33.7 & 87.7  \\
    Deep Ensemble(2017)  & 7.2 & 25.4 & 90.0    \\
    AutoEncoder(2018)  & 2.2 & 91.7 & 66.1 \\
    MSP+CRF(2019)  & 6.5 & 29.9  &  88.1   \\
    TRADI(2020)  & 7.2 & 25.3  & 89.2   \\
	SynthCP(2020)  & \color{blue}{\textbf{9.3}} & 28.4  & 88.5   \\
	PAnS(2021)  & 8.8 & \textcolor{blue}{\textbf{23.2}}  & \textcolor{blue}{\textbf{91.1}}   \\
	\hline 
	$\textbf{DiCNet}$   & \color{red}{\textbf{15.6}} & \color{red}{\textbf{18.0}} & \color{red}{\textbf{93.3}}\\
	\hline 
    \end{tabular}
\caption{Anomaly discovery results on StreetHazards dataset \cite{Hendrycks02}.}
\label{tab:result}
\end{table}

\subsection{Quantitative Results}
Tab. \ref{tab:result} shows the quantitative comparisons on the StreetHazards dataset. 
DiCNet outperforms all other methods in three metrics.
In particular, in FPR95, DiCNet achieves an improvement of 5.2\% (From 23.2\% to 18.0\%), 
which indicates that when the recall is fixed at 95\%, 
DiCNet has a lower false positive rate than existing approaches.
As previously mentioned in Tab. \ref{tab:efpr}, 
the key improvement is induced by reducing the C-FPR and E-FPR.
DiCNet improves the previous state-of-the-art approach SynthCP from 9.3\% to 15.6\% in terms of AUPR.
For AUROC, DiCNet also achieves an improvement of 2.2\% compared with existing methods.
Tab. \ref{tab:result_bdd} shows the quantitative comparisons between our method and prior methods on the BDD-Anomaly dataset. 
As is known, BDD-Anomaly is a challenging dataset since many anomaly objects have a small size.
Even under such circumstances, 
DiCNet achieves a 6.8\% improvement in FPR95 and 4.2\% improvement in AUPR,
which significantly outperform other state-of-the-art approaches.

\begin{table}[tb]
\small
\centering
    \begin{tabular}{l|c|c|c}
    \hline
	Method   & AUPR $\uparrow$ & FPR95 $\downarrow$  & AUROC $\uparrow$    \\  
	\hline
	\hline
	MC Dropout(2016)   & 6.5 & 33.4 & 83.7    \\
	MSP(2017)   & 6.3 &  31.9  & 84.2\\
	Deep Ensemble(2017)   & 6.0  & \color{blue}{\textbf{25.0}}   & \color{blue}{\textbf{87.0}} \\
	AutoEncoder(2018)   & 1.4 & 82.2 & 56.8 \\
	MSP+CRF(2019)  & \color{blue}{\textbf{8.2}} & 26.0  &  86.3   \\
	TRADI(2020)     & 5.6 & 26.9  & 86.1    \\
	\hline 
	$\textbf{DiCNet}$  & \color{red}{\textbf{12.4}} & \color{red}{\textbf{18.2}} & \color{red}{\textbf{91.9}}\\
	\hline 
    \end{tabular}
\caption{Anomaly discovery results on BDD-Anomaly dataset \cite{Hendrycks02}.}
\label{tab:result_bdd}
\end{table}

\subsection{Qualitative Analysis}

Fig. \ref{fig:compare} illustrates the qualitative comparisons between 
MSP \cite{Hendrycks19}, SynthCP \cite{Xia03} and DiCNet. 
Given the input image shown in Fig. \ref{fig:compare} (a), 
MSP, SynthCP and DiCNet use PSPNet \cite{Zhao05} to predict the semantic class of each pixel, 
which produces the results shown in Fig. \ref{fig:compare} (c).
Compared with the ground truth in Fig. \ref{fig:compare} (b),
the pixels around the edges are easily incorrectly classified,
e.g., the pixels near the edge of a tree.
As seen in Fig. \ref{fig:compare} (e),  
MSP tends to assign a higher anomaly score to the pixels near the edges.
Similarly, in Fig. \ref{fig:compare} (f), 
SynthCP also reports some anomalies near the edges. 
These visualizations clearly reflect that both MSP and SynthCP suffer from the presence of incorrectly classified pixels. 
Such a phenomenon corresponds to a higher E-FPR scores reported in Tab. \ref{tab:efpr}.
In addition, some correct classified pixels are still assigned a high anomaly score by MSP and SynthCP.
Such a phenomenon can also be observed in the C-FPR scores in Tab. \ref{tab:efpr}.
In contrast, as we observed in Fig. \ref{fig:compare} (g), 
DiCNet assigns a much lower anomaly score to the incorrectly classified pixels, e.g., the pixels around the edges, 
which corresponds to a lower false positive rate given in Tab. \ref{tab:efpr} and Tab. \ref{tab:result}.
Such results intuitively demonstrate that DiCNet is capable of effectively alleviating the essential problem caused by incorrect semantic classification.
In addition, in the correct classified regions, DiCNet produces fewer anomalies. 
Such results reflect the reason for the improvement in C-FPR.
Furthermore, when we compare the anomaly maps in the first row with the ground truth shown in Fig. \ref{fig:compare} (d), 
we find that DiCNet also reduces the false negative rate, 
e.g., the wheels of the yellow muck truck are regarded as normal by MSP and SynthCP, 
while DiCNet predicts the correct anomaly score for the wheels.

\begin{table}[tb]
\footnotesize
\centering
\begin{tabular}{c|c|c|c|c|c}
    \hline
	\scriptsize{Backbone}   &\scriptsize{Epoch} & \scriptsize{AUPR$\uparrow$} & \scriptsize{FPR95 $\downarrow$}   & \scriptsize{{}AUROC$\uparrow$}  &\scriptsize{Time}\\  
	\hline
	\hline
	\multirow{4}{*}{ResNet-101} 
	&10 & 11.6 & 27.5 & 88.3 & \multirow{4}{*}{450.2ms}  \\
	&25 & \textbf{\color{red}15.8} & 18.7 & 93.1   \\
	&50 & 15.6 & 20.8 & 92.2   \\
	&110 & 14.7 & 24.0 & 90.9   \\
    
	\hline
	\multirow{6}{*}{ResNet-50} 
	&3 & 12.5 & 19.7  & 92.5 & \multirow{6}{*}{347.8ms}  \\
	&4 & 14.1 & 20.7  & 92.3   \\
	&5 & 15.4 & 20.4  & 92.4   \\
	&10 & 14.2 & 22.3  & 91.5   \\
	&50 & 14.9 & 26.1  & 89.9   \\
	&110 & 12.2 & 28.3  & 88.6   \\
	\hline
	\multirow{4}{*}{ResNet-34} 
	&10 & 10.6 & 18.7  & 92.5 & \multirow{4}{*}{303.6ms} \\
	&50 & 13.8 & \textbf{\color{blue}{18.1}}  & 93.1  \\
	&100 & 15.2 &18.5  & 92.9  \\
	&110 & 15.6 & \textbf{\color{red}{18.0}}  & \textbf{\color{red}93.3}  \\
	\hline
	\multirow{4}{*}{ResNet-18} 
	&10 & 11.5 & 19.0 & 92.5 & \multirow{4}{*}{261.7ms} \\
	&50 & 14.4 & 18.8 & 93.0 \\
	&100 & \textbf{\color{red}15.8} & 18.2 & \textbf{\color{blue}93.2}\\
	&110 & \textbf{\color{blue}15.7} & 18.5 & 93.1 \\
	\hline
	\multirow{4}{*}{VGG-19} 
	&20   & 8.9 & 23.5 & 90.4  & \multirow{4}{*}{242.5ms}  \\
	&50   & 11.8 & 22.9 & 90.9    \\
	&80   & 14.1 & 21.1 & 92.0    \\
	&110   & 13.9 & 21.6 & 91.7    \\
	\hline
	\multirow{4}{*}{AlexNet} 
	&20 & 9.2 & 23.3 & 90.8 & \multirow{4}{*}{236.4ms} \\
	&50 & 10.1 & 22.9 & 91.1    \\
	&80 & 10.6 & 22.7 & 91.2    \\
	&110 & 10.8 & 22.7 & 91.3    \\
	\hline 
\end{tabular}
\caption{Results of different backbones for student branch on StreetHazards.
\emph{Note that `Time' in the last column indicates the total inference time of DiCNet, not the student branch.}}
\label{tab:backbone}
\end{table}

\subsection{Discussions}
\label{sec:discussions}

In this section, we report several variations of DiCNet and discuss their influences on anomaly discovery. 
\yz{Note that, all the following experiments adopt ResNet-50 based PSPNet (without the semantic classification head) as the teacher branch}.

\noindent \textbf{Effect of Model Size.} 
As we discussed in Sec. \ref{sec:distillation}, 
DiCNet is flexible in architecture design,
we can distill the knowledge of the fixed teacher branch to any model size.
To verify such a claim, we conduct experiments to study the effect of different model sizes of the student branch.
Several key results are reported in Tab. \ref{tab:backbone}.
As we can see, the accuracy of different ResNet based backbone is closed to each other.
e.g., ResNet-101 and ResNet-18 both obtain an AUPR score of 15.8.
ResNet-34 achieves the best results in both FPR95 and AUROC.
ResNet-101, ResNet-50 and ResNet-18 can also produce comparable results in these two metrics.
However, a key mechanism of DiCNet is that different backbone has a different convergence speed. 
As seen in Tab. \ref{tab:backbone}, ResNet-50 achieves its best results \yz{at epoch 5}.
Similarly, the best results of ResNet-101 is acquired \yz{at epoch 25.}
\yz{In contrast}, if the backbone of the student branch is smaller than the teacher branch, 
more training epochs are demanded to ensure the consistency of the teacher branch and the student branch, 
e.g., for ResNet-34 and ResNet-18, we need more than 100 epochs to achieve their best results.
Similar phenomenon can also be observed in VGG-19 and AlexNet.
In addition, Tab. \ref{tab:backbone} also reported the inference time of DiCNet.
By jointly considering the accuracy, the convergence speed and the inference time, 
we report the ResNet-34 based DiCNet as the best trade-off result.

\begin{figure}[t]
\centering
\includegraphics[width=1.0\columnwidth]{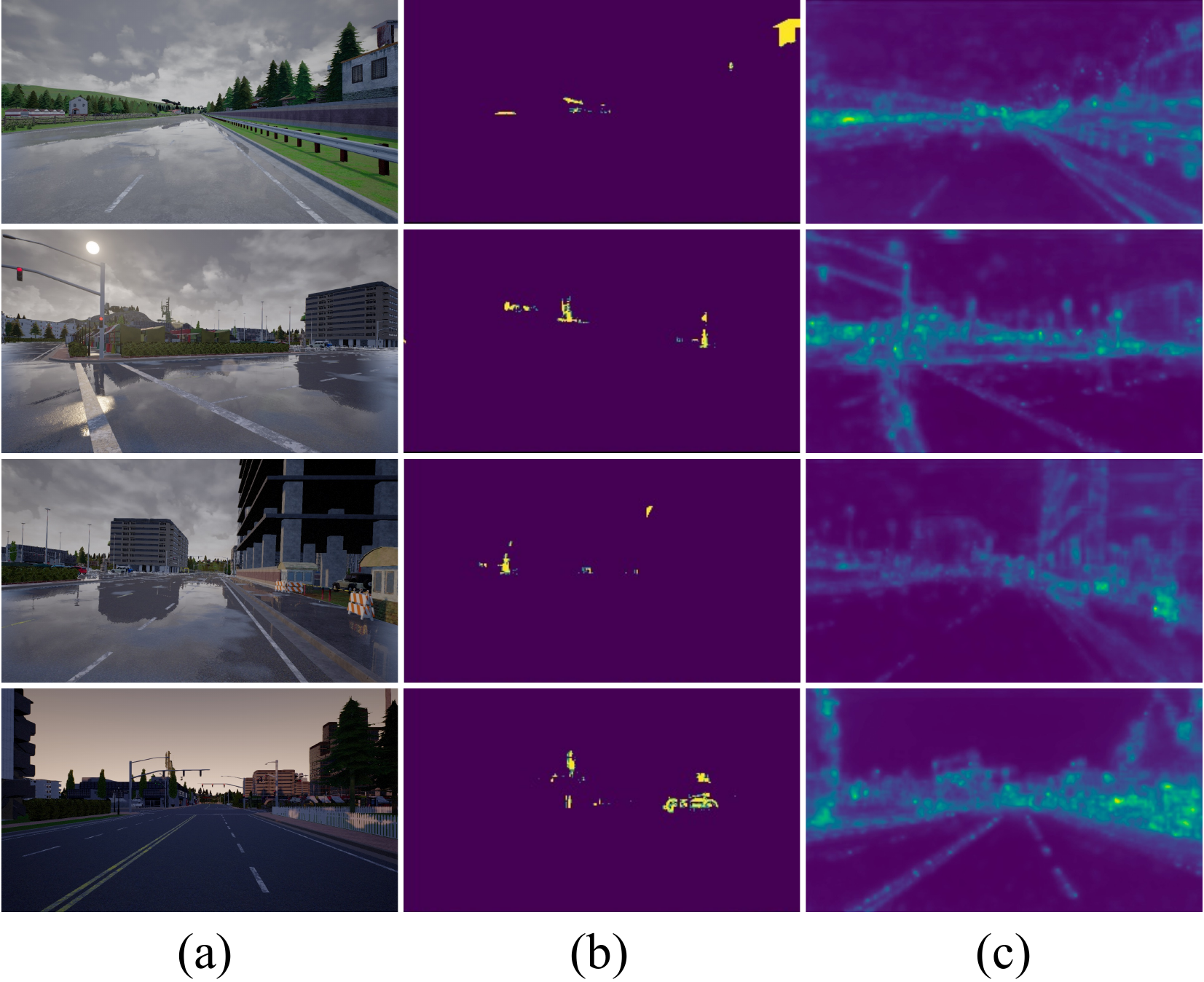} 
\caption{Some failure examples. (a) the input images, (b) ground truth of anomaly, (c) the anomaly score maps. Best view in color.}
\label{fig:bad}
\end{figure} 

\subsection{Failure Cases Analysis}
As demonstrated in previous experiments, 
DiCNet performs well in most cases of detecting anomalies. 
However, it still fails in several challenging cases, 
mainly including serious occlusion and small objects. 
Some failure examples are given in Fig. \ref{fig:bad}. 
DiCNet also has some false positives on some rarely seen normal objects. 
Note that all these errors are common challenges for the other state-of-the-art methods.
We will try to address these issues in our further works.

\section{Conclusion}
\yz{
This paper proposes a novel Distillation Comparison Network (DiCNet) for anomaly discovery in semantic segmentation,  
which contains a teacher branch and a student branch. 
The former one is trained under the common setting of semantic segmentation, 
the latter one is distilled from the fixed teacher branch through a distribution distillation.
We leverage the discrepancy of the semantic feature between the two branches to discover the anomalies. 
DiCNet alleviates the issue caused by incorrect semantic classification efficiently. 
We conduct extensive experiments on two challenging datasets to verify the effectiveness of DiCNet. 
}

\bibliography{aaai22.bib}

\end{document}